%% file: root.tex
\documentclass[letterpaper, 10 pt, conference]{ieeeconf}  % Comment this line out if you need a4paper

\IEEEoverridecommandlockouts
\usepackage{graphics} % for pdf, bitmapped graphics files
\usepackage{amsmath} % assumes amsmath package installed
\usepackage[pdftex]{graphicx}
\graphicspath{{figures/}}
\usepackage{hyperref}
\hypersetup{
	colorlinks=true,
	linkcolor=black,
	citecolor=black,
	filecolor=black,
	urlcolor=black,
}
\usepackage{verbatim}   %use this package to comment multiline 
\usepackage[T1]{fontenc}
\usepackage[utf8]{inputenc}
\usepackage{csquotes}
\usepackage[english]{babel}
\usepackage[export]{adjustbox}
\usepackage{caption}
\usepackage{subcaption}
\usepackage[colorinlistoftodos]{todonotes}
\usepackage{lipsum}
\usepackage{amsmath} 

\usepackage{mathtools}
\usepackage{amssymb}
\usepackage{placeins}%for \FloatBarrier
\usepackage{pgfplots}
\pgfplotsset{compat=1.11}
\usepackage{algorithm}
\usepackage{algorithmic}
\usepackage{array,booktabs}
\usepackage{tikz}
\usepackage{multirow,graphicx}
\usepackage{makecell}
\usetikzlibrary{matrix,decorations.pathreplacing}
\usepackage[style=ieee,
doi=false,
url=false,
mincitenames=1,
maxcitenames=1,
minbibnames=6,
maxbibnames=6,
backend=bibtex]{biblatex}  % supports bibliography with BibLaTeX
\input{misc/commands.tex}
\title{%
High-level Decisions from a Safe Maneuver Catalog with Reinforcement Learning for Safe and Cooperative Automated Merging 
}

\author{Danial Kamran$^{1}$, Yu Ren$^{2}$ and Martin Lauer$^{1}$% <-this % stops a space
	\thanks{$^{1}$Authors are with Institute of Measurement and Control Systems, Karlsruhe Institute of Technology (KIT), 76133 Karlsruhe, Germany, {\tt\small \{danial.kamran, martin.lauer\}@kit.edu }}%
	\thanks{$^{2}$Yu Ren is student at  Karlsruhe Institute of Technology (KIT), 76133 Karlsruhe, Germany, {\tt\small yu.ren@student.kit.edu}}%
}

\begin{document}
\maketitle

%% add copyright notice for arxiv or other non-IEEE publications, see https://ieeeauthorcenter.ieee.org/publish-with-ieee/author-education-resources/guidelines-and-policies/policy-posting-your-article/
%% 1. enable \IEEEoverridecommandlockouts
%% 2. uncomment \IEEEpubidadjcol and the appropriate \IEEEpubid section
%% 3. place \IEEEpubidadjcol also in the second column of the title page (probably somewhere in your introduction section)
%% 4. disable the \thispagestyle{empty} command
%% Use this after submission for review (but not after acceptance):
% \IEEEpubid{\begin{minipage}{\textwidth}~\\[12pt] \centering%
%   This work has been submitted to the IEEE for possible publication. Copyright may be transferred without notice, after which this version may no longer be accessible.
% \end{minipage}}
% Use this after acceptance:
\pubid{\begin{minipage}{\textwidth}~\\[12pt] \centering%
		%   10.1000/xyz123~ % Insert your DOI after publication
		\copyright~2021 IEEE. Personal use of this material is permitted. Permission from IEEE must be obtained for all other uses, in any current or future media, including reprinting/republishing this material for advertising or promotional purposes, creating new collective works, for resale or redistribution to servers or lists, or reuse of any copyrighted component of this work in other works.
\end{minipage}}
\pubidadjcol

\pagestyle{empty}

\input{content/01_abstract}
\input{content/02_introduction}
\input{content/03_main}

\input{content/04_results_and_evaluation}

\input{content/05_conclusions}
\input{06_acknowledgements}
\printbibliography

\end{document}

%% file: misc/commands.tex
\renewcommand{\S}{\mathcal{S}}
\newcommand{\A}{\mathcal{A}}
\renewcommand{\O}{\mathcal{O}}

\newcommand{\tm}{\ensuremath{P}}

\newcommand{\om}{\ensuremath{\mathcal{Z}}}

\renewcommand{\rm}{\ensuremath{R}}

\newcommand{\mpsc}{\frac{\text{m}}{\text{s}^3}}
\newcommand{\mpsq}{\frac{\text{m}}{\text{s}^2}}
\newcommand{\mps}{\frac{\text{m}}{\text{s}}}

%% file: content/01_abstract.tex
% !TeX root = ../root.tex
% -*- root: ../root.tex -*-
\begin{abstract}
Reinforcement learning (RL) has recently been used for solving challenging decision-making problems in the context of automated driving.
However, one of the main drawbacks of the presented RL-based policies is the lack of safety guarantees, since they strive to reduce the expected number of collisions but still tolerate them.
In this paper, we propose an efficient RL-based decision-making pipeline for safe and cooperative automated driving in merging scenarios.
The RL agent is able to predict the current situation and provide high-level decisions, specifying the operation mode of the low level planner which is responsible for safety.
In order to learn a more generic policy, we propose a scalable RL architecture for the merging scenario that is not sensitive to changes in the environment configurations.
According to our experiments, the proposed RL agent can efficiently identify cooperative drivers from their vehicle state history and generate interactive maneuvers, resulting in faster and more comfortable automated driving.
At the same time, thanks to the safety constraints inside the planner, all of the maneuvers are collision free and safe.
\end{abstract}

%% file: content/02_introduction.tex
\section{Introduction}

\begin{figure}
	\centering
	\includegraphics[width=1.\linewidth]
	{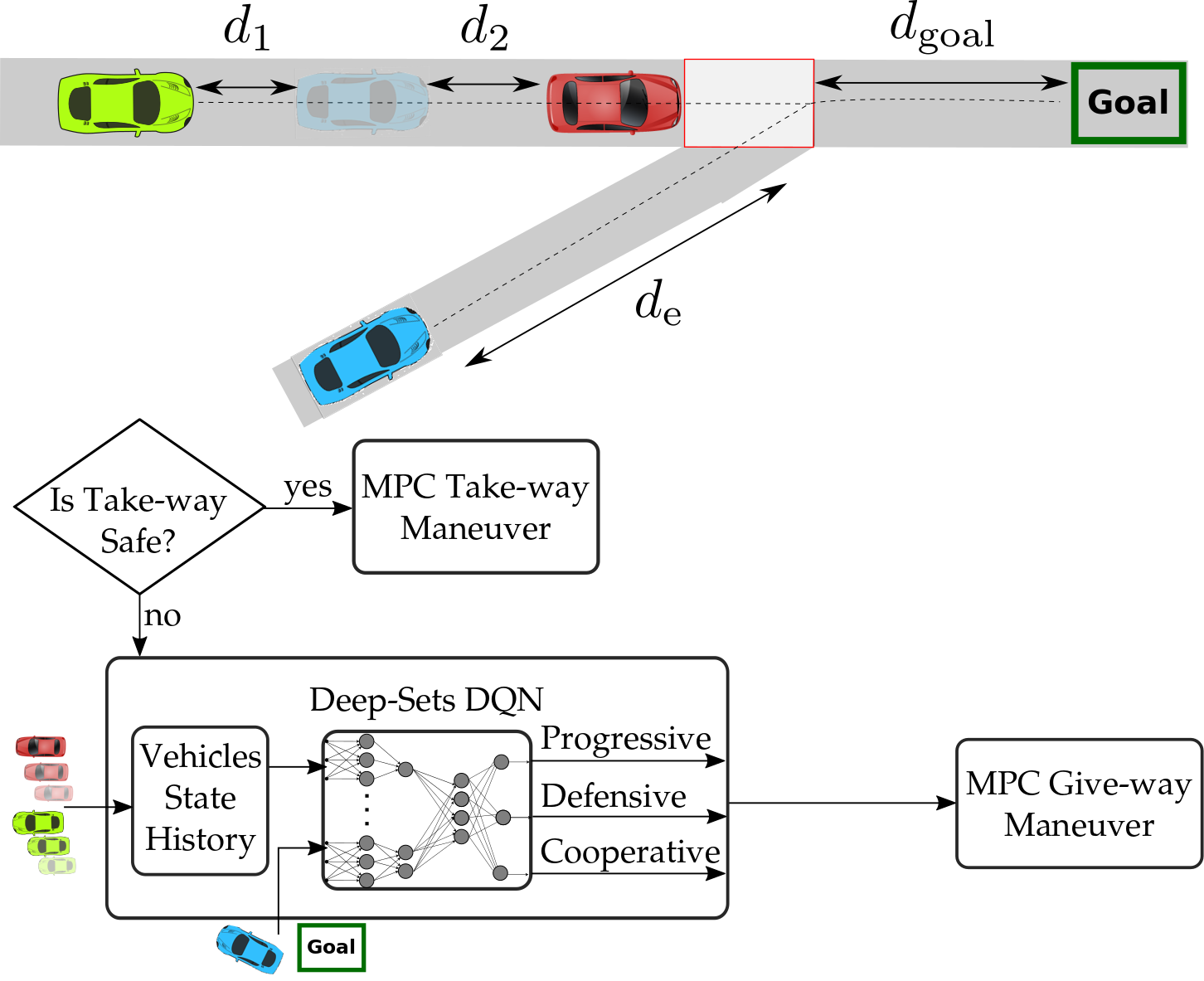}
	\caption{Merging scenario and the proposed Deep-Sets DQN pipeline to provide safe and cooperative maneuvers. Some vehicles (green vehicle) cooperatively reduce their velocity to open a merging gap for the blue vehicle which is ego vehicle. }
\label{fig:first_page}
\end{figure}

%Here we explain the main challenges and motivations for this paper.
%Then we explain some related papers and finally we explain the main contributions in one paragraph and paper structure in another paragraph
One of the most important challenges for automated driving in urban environments is providing safe and cooperative policies that can efficiently predict the future situation and generate optimal decisions based on that.
An optimal decision for the automated vehicle requires precise information about intention of other participants which is not completely achievable by the existing environment perception approaches.
Specifically for merging scenarios, other participants can have different reactions with respect to the ego vehicle.
Some of the drivers may be cooperative and reduce their velocity to open a merging gap for the ego vehicle, while others may be aggressive and even close the existing gap.
On the other hand, some drivers may be distracted and do not react to the ego vehicle at all (cf. Figure \ref{fig:first_page}).
Therefore, the efficiency of the automated vehicle behavior highly depends on the accuracy of other drivers intention estimation and their behavior prediction.

Reinforcement Learning (RL) has recently gained a huge attention in solving complex robotic tasks specifically for efficient automated driving \cite{isele_navigating,bouton2019cooperation,tram2019rl_mpc, safe_multi_agent, kamran2020risk}.
The main advantage of RL based decision-making agents is their ability to learn long term optimal actions by experiencing multiple interactions with the environment.
Some approaches used RL in order to learn optimal policies for merging scenarios. 
Tram et. al in \cite{tram2019rl_mpc} proposed to combine RL with model predictive control (MPC) in order to solve negotiation problem in merging scenario. 
They provide distances and velocities of surrounding vehicles as the input of recurrent Deep Q Network (DQN) that selects which vehicle should be followed through setting appropriate constraints inside MPC planner.
In another work, Bouton et. al in \cite{bouton2019cooperation} proposed to estimate a belief over other drivers cooperation level which is provided as the input of an RL agent that decides about acceleration of the ego vehicle. 
The belief estimator, however, was not part of the learning framework and could be included implicitly inside the RL which is one of the contributions in this paper.

\pubidadjcol

An important challenge is providing safe and efficient policies based on RL that are not only safe in majority of experiments but are reliable to be used as part of the automated driving pipeline in reality.
In most of the previous works using RL like \cite{tram2019rl_mpc, bouton2019cooperation, isele_navigating, kamran2020risk}, the agent is punished by a negative reward for being in risky or collision states in order to encourage learning safer policies.
Although this approach can help to reduce number of failures and risky behaviors during training, still there is no guarantee for anytime safety.

There are some approaches which are not based on RL and implement safety as hard constraints inside an optimization framework to provide safe trajectory planning.
In an optimization based approach, M{\"u}ller et. al in \cite{muller2019risk} suggested to predict future trajectories of other drivers using a Kalman filter with underlying constant velocity model and apply those predictions into an optimization problem which optimizes comfort of the maneuvers.
However, constant velocity assumption for other vehicles future trajectories prohibits these approaches to be used in more realistic scenarios where other drivers may aggressively accelerate to close a merging gap or cooperatively decelerate to open a merging gap for the ego vehicle.

Using reachability analysis \cite{althoff2014online}, some approaches guarantee safety considering all reachable areas of other vehicles and make sure that the ego vehicle never enters into areas that can be occupied by other vehicles regardless of their intention and future trajectory predictions.
Although these approaches can guarantee safety, the main drawback is their lack of ability to interact with other participants which results in conservative behaviors without considering reaction to cooperation signals from other drivers or performing \textit{information gathering} actions to identify cooperative drivers.

Considering uncertain behavior of other participants as a partial observation, some researchers formulated the cooperative merging scenario as a Partially Observable Markov Decision Process (POMDP).
Hubmann et al. in \cite{hubmann2018merging} proposed combining a POMDP solver with a trajectory planner to generate interactive maneuvers which can identify suitable merging gaps in a dense traffic scenario.
Although these approaches can tackle all requirements for a cooperative behavior in these scenarios, they have challenges regarding processing time and scalability.

In this paper, we propose to combine RL with an optimization based trajectory planner to provide safe and interactive trajectories.
Similar to \cite{hubmann2018merging} we formulate the merging problem as a POMDP where intention of other participants are hidden.
However, in order to find the optimal behavior policy, we propose a  scalable Deep Q Network (DQN) as an RL agent which implicitly predicts others intention and selects the best action. 
The proposed scalable DQN agent can handle complex situations with multiple number of vehicles and various cooperation levels which is challenging to be real-time for conventional POMDP solvers.
Moreover, since the safety constraints are applied inside the trajectory planner, the proposed RL approach can provide safety guarantees, which is not the case for similar RL-based approaches like (\cite{tram2019rl_mpc, bouton2019cooperation, isele_navigating, kamran2020risk}).

In the remaining parts of this paper, we first describe our POMDP formulation for automated driving at merging scenario in section \ref{sec:problem_formulation}. 
Then we introduce the proposed safe and cooperative decision-making pipeline for merging scenarios in section \ref{sec:approach}.
Finally we will express the simulation environment and evaluation results in section \ref{sec:evaluations} and have conclusions in section \ref{sec:conclusions}.

%Proposing a multi-model MPC planner, we generate safe trajectories that are operating in different modes regulated by a high %level learning based agent.
%In this framework, RL as the learning-based decision making module provides high level actions for MPC-based trajectory planner for safe and cooperative automated driving.
%The RL-agent implicitly predicts the behavior and intention of other drivers and provides optimal actions which are used in the low-level trajectory planner which is responsible for safety and feasibility of generated trajectories.

%The main contributions of this paper are:
%\begin{itemize}
%\item 	We apply memory-based RL to solve a POMDP framework for merging scenario where the intention of other drivers are hidden and are implicitly predicted using history of observations provided as the input of RL agent.
%
%
%\item We provide anytime safety verification in order to guarantee safety of generated trajectories even if the RL fails to correctly predict other drivers intention.
%
%\item An scalable DQN network based on Deep-sets structure \cite{zaheer2017deepsets, huegle2019dynamic} is proposed for merging scenarios which helps to learn generic policies that are capable to handle merging in environments with different traffic densities.
%\end{itemize}

\section{Problem Formulation}\label{sec:problem_formulation}
The main challenge for automated merging scenarios is due to the uncertainty about intention of other drivers.
We assume some of them are \textit{cooperative} trying to open a merging gap for easing ego vehicle merge while some others are \textit{distracted} and do not react to the ego vehicle.
Therefore, we consider automated merging scenario as a partially observable Markov Decision Process (POMDP) $(\S, \A, \O, \tm, \om, \rm, \gamma)$, where $\S$ is the state space, $\A$ the action space, $\O$ the observation space, $\tm$ the transition function, $\rm$ the reward function and $\gamma \in [0,1)$ the discount factor.

Similar to \cite{tram2019rl_mpc}, we find the best policy which maximizes the future expected cumulative reward (return) as $\sum_{t=0}^{\infty} \gamma^t r_t$ using reinforcement learning \cite{sutton_barto_rl}.
Due to partial observation, at each discrete time step $t$, the RL agent receives an observation $o_t \in \O$ according to the current state $s_t$ and the observation model $\om$ and chooses an action based on the policy $\pi$.
The distribution of successor state $s_{t+1}$ is defined by the transition model $s_{t+1}\sim\tm(s_t,a_t)$ based on the current state and chosen action.

%%%%%%%%%%%%%%%%%%%%%%%%%%%READ AND IMPROVE THIS PART AT LEAST FOR CAMERA READY VERSION OR EARLIER%%%%%%%%%%%%%%%%%%%%%%%%%%%
\subsection{Deep $Q$-Learning}
A popular off-policy algorithm for solving the reinforcement learning problem is $Q$-learning
\cite{qlearning}.
Instead of directly optimizing the policy, this algorithm tries to find
the value function $Q_\pi$ of the optimal policy.
The value function
\begin{equation*} 
\label{eq:q}
Q^\pi(s_t,a_t) = \mathbb{E}_{s_i\sim\tm}\left[ \rm(s_t,a_t) + \sum_{k=1}^\infty \gamma^{k} \rm(s_{t+k}, \pi(s_{t+k})) \right], 
\end{equation*}
is defined as the expected return when choosing action $a_t$ in state $s_t$ and following policy $\pi$ thereafter \cite{sutton_barto_rl}.
The main goal in RL is to maximize the return.
Using the Bellman equation \cite{bellman_dynamic}, the optimal value function can be represented as:
\begin{equation*}
Q^*(s_t,a_t) = \mathbb{E}_{s_i\sim\tm}\left[ \rm(s_t,a_t) + \gamma \max_{a^\prime} Q^*(s_{t+1}, a^\prime)) \right].
\end{equation*}

In Deep Q networks (DQN) \cite{dqn} the optimal value function is estimated by a neural network with parameters $\theta$ which is iteratively trained by minimizing the squared temporal difference (TD) error:
\begin{equation*}
\delta^2_t = \left[ r_t + \gamma \max_{a^\prime} Q_\theta(s_{t+1}, a^\prime) -  Q_\theta(s_t,a_t)\right]^2,
\end{equation*}
over samples ($s_t, a_t, r_t, s_{t+1}$) from a replay buffer recording transitions of an $\epsilon$-greedy policy over $Q_\theta$, i.e. selecting action with maximum $Q_\theta$ with 1-$\epsilon$ probability and a uniformly random action otherwise.   

%% file: content/03_main.tex
\section{Proposed Approach}
\label{sec:approach}

In the merging scenario, the ego vehicle should prevent entering into the conflict zone when there are other vehicles driving close to it. 
However, some of the vehicles may reduce their velocity to open a merging gap for the ego vehicle and some others may behave aggressively to close the existing merging gap or drive with constant velocity.
Our approach focuses on both safe and cooperative factors for automated driving in merging scenarios.
For that, we utilize an RL agent to implicitly identify cooperative behavior of other drivers and decide about suitable maneuvers from a safe \textit{maneuver catalog} which are provably safe according to the worst-case predictions of other vehicles inside the low level planner (cf. Figure \ref{fig:first_page}).

\subsection{Safe Trajectory Planning with Model Predictive Control}
\begin{figure*}[t]\centering
	\fontsize{7.5pt}{11pt}\selectfont
	\def\svgwidth{0.99\linewidth}
	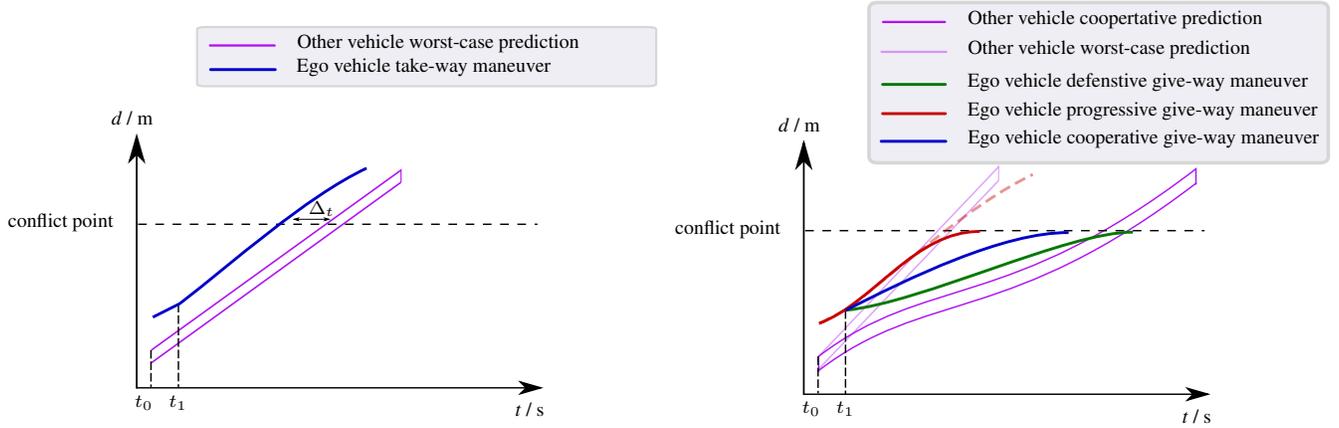
	\caption{\textbf{Left:} Generated take-way maneuver which is safe according to the worst-case prediction for other vehicle (driving with $v_\text{max}$). \textbf{Right:} Situation where take-way maneuver is not safe considering the worst-case prediction of other vehicle and RL agent decides which of the three possible give-way maneuvers (red, blue or green) for the ego vehicle should be generated.}\label{fig:merging_mpc_options}
\end{figure*}

In order to generate safe and comfortable maneuvers, we rely on Model Predictive Control (MPC) for efficient longitudinal trajectory planning in merging scenario which considers safety as \textit{hard} constraints and comfort as \textit{soft} constraints inside the optimization framework.
We also consider the trajectory planning only in longitudinal direction which is usable for most of the merging situations such as merging at unsignalized intersections or in highways.

Therefore, in the proposed MPC planner we define vehicle state $\mathbf{x} \in \mathbb{R}^3$ and control $\mathbf{u} \in \mathbb{R}$ as:
\begin{equation}
\mathbf{x_{k}} = \left(
\begin{array}{cccc}
d_{k}, v_{k}, a_{k} 
\end{array}\right),
\end{equation}
where $\mathbf{x_{k}}$ represents the MPC state in time step $t_k$ and  $d_k$, $v_k$, $a_k$ represent the ego vehicle longitudinal distance along road, its velocity and its acceleration respectively. 

Similar to \cite{muller2019risk}, we optimize the MPC cost over jerk commands to directly control the amount of applied jerk considering passengers comfort:
\begin{equation}
\mathbf{u_{k}} = j_{k},
\end{equation}
where $j_k$ is the jerk command being optimized for the optimization step $k$ by MPC. 
Moreover, ego vehicle follows the following dynamics:
\begin{equation}
\mathbf{d}_{k+1} = \mathbf{d}_k + \mathbf{v}_k\Delta{t} + \frac{1}{2}\mathbf{a}_k\Delta{t}^2
\end{equation}

The main goal is to drive fast and comfortable if no safety constraint is violated. 
Therefore, the problem can be regarded as a finite horizon, constrained optimal control problem with this quadratic cost function:
\begin{align}
J =\sum_{i=0}^{N} (\mathbf{x}_k-\mathbf{x}^{ref}_k)^\mathrm{T}  Q(\mathbf{x}_k -\mathbf{x}^{ref}_k)  + c(\mathbf{u}_k)
\end{align}
\begin{align}
{\large {s.t.}} \qquad &\mathbf{x}_{k+1}  = A \mathbf{x}_k + B \mathbf{u}_k \\
&h(\mathbf{x}_k, \mathbf{u}_k , \mathbf{o})  \le 0 
\end{align}
%here we can use set the detailed dynamics of ego vehicle.
\begin{equation}
\mathbf{x}_{k+1}={
	\left[ \begin{array}{ccc}
	1 & \Delta{t} & \frac{1}{2}\Delta{t}^2\\
	0 & 1 & \Delta{t}\\
	0 & 0 & 1
	\end{array} 
	\right ]}\mathbf{x}_k + {
	\left[ \begin{array}{c}
	\frac{1}{6}\Delta{t}^3 \\
	\frac{1}{2} \Delta{t}^2 \\
	\Delta{t} \\
	\end{array} 
	\right ]}\mathbf{u}_k
\end{equation}

where $Q$ is weight matrix, $\mathbf{x}_{ref}$ the reference trajectory for driving with $v_\text{ref}$ along the road and $N$ is the number of optimization steps.
In addition, $c(.)$ is the cost for the jerk, specifying how big the emergency braking is important which is different for each of maneuvers in our maneuver catalog providing multiple safe maneuver choices.
Moreover, constraint (4) enforces the system dynamic and constraint (5) is the general form of all constraints for initial state, boundary of state and control input, as well as the safety constraint for different sub-maneuvers which will be explained later.

In contrast to similar approaches like \cite{tram2019rl_mpc} that assume constant velocity for other vehicles future trajectory prediction or \cite{batkovic2019mpc_prediction} which rely on external prediction modules, we consider the worst-case predictions for safety constraints inside MPC and provide anytime safe trajectories which are summarized here:
\begin{itemize}
	\item On the merging lane, the closest vehicle behind the conflict area accelerates with $a_{\text{max}}$=$4\mpsq$  to reach velocity of $v_{\text{max}}$=$15\mps$.
	\item On the merging lane, the closest vehicle after the conflict area decelerates with $a_{\text{min}}$=$-4\mpsq$  to reach zero velocity.
\end{itemize}

Figure \ref{fig:merging_mpc_options} shows an example for the worst case prediction of other vehicles in a merging scenario considered in our planner.

The proposed MPC planner operates in two major modes: safe take-way maneuver and safe give-way maneuver.
At every planing step $t$, the MPC planner receives the current observation $o_t$ and first verifies if a safe take-way maneuver considering the worst-case prediction for other vehicles is feasible or not (cf. Figure \ref{fig:merging_mpc_options} left).
In case the safe take-way maneuver is not feasible and the MPC planner can not find a solution for that, we consider the safe give-way maneuver which should be always feasible when the ego vehicle is behind the conflict zone as a fail-safe plan.

\subsubsection{Safe Take-way Maneuver}
In this maneuver, the ego vehicle plans to enter into the merging area and leave there $\Delta_{\text{t}}$=$0.5 \text{s}$ earlier than the time the closest vehicle behind or inside the conflict zone in the worst case can enter there. 
It also prevents to have a collision with the closest vehicle after the conflict zone by assessing a full stop maneuver with $\Delta_{\text{d}}$=$0.5 \text{m}$ safety distance is feasible. 
Therefore, the safety constraints for this maneuver are defined as:

\begin{equation}
d_{\lfloor (t_\text{c}-\Delta_\text{t})/t_s \rfloor } \ge d_\text{ce},
\end{equation}
\begin{equation}
d_N \le d_\text{max}-\Delta_{\text{d}}, v_N = 0,
\end{equation}
where $t_\text{c}$ is the worst case time the closest vehicle before the conflict zone can enter there.
$t_\text{c}$ is set to -1 when the vehicle is already inside the conflict zone (take-way is not feasible).
$t_s$ is the MPC time step, $d_\text{ce}$ the relative distance of the conflict zone ending edge, and $d_\text{max}$ is the relative distance of a possible front vehicle on the route.
For comfortable driving, we add the comfort cost function for this maneuver which is described in table \ref{tab:give_ways_costs}.

\subsubsection{Safe Give-way Maneuver}
For this maneuver, the safety constraint is formulated in a way to make sure that the ego vehicle stops in front of the conflict zone:
\begin{equation} 
d_N \le d_{cs}, v_N = 0,
\label{eq:give_way_constraint}
\end{equation}
where $d_{cs}$ is the longitudinal distance of the conflict zone starting edge on the ego vehicle route. 
For the comfort cost function of the give-way maneuver, we define three different sub-maneuvers with different cost functions: \textit{Progressive Give-way}, \textit{Defensive Give-way} and \textit{Cooperative Give-way}.
The main goal to define these three sub-maneuvers is to enable interaction with other vehicles and formulate information gathering and decision postponing behaviors decided by the RL agent as the high level decision-making module.
All of these maneuvers have same safety constraints defined in equation \ref{eq:give_way_constraint}, but they have different cost functions (shown in Table \ref{tab:give_ways_costs}) which effect the amount of conservativeness and comfort.
The goal of progressive mode is to drive faster to increase the chance of a possible take-way maneuver in the future, thus the negative jerk commands for the second half of trajectory is punished with small weight ($\text{W}_\text{L}$).
The aim of Defensive mode is to comfortably reduce the velocity and stop, thus it punishes the negative jerk in the whole trajectory with a relatively big weight ($\text{W}_\text{H}$). 
The \textit{Cooperative} sub-maneuver is selected when it is not clear which of the above two sub-maneuvers to choose and drives with constant velocity in order to decide for a suitable sub-maneuver in the next decision time.  
Therefore, it punishes both positive and negative jerk commands with $W_N$.
In our experiments we set $\text{W}_\text{N}=0.5$, $\text{W}_\text{L}=0.005$, $\text{W}_\text{H}=5000$ and $\text{W}_\text{C}=1.0$ to generate proper maneuvers for each mode using the MPC planner.
\begin{table}[h]
	\centering
	\caption{Comfort cost function for different modes}\label{tab:give_ways_costs}
	\begin{tabular}{c|c}
		\hline
		safe take-way  &  $c(\mathbf{u}_k) = W_N \cdot (\mathbf{u}_k)^2$   \\\hline
		Progressive mode  &  $c(\mathbf{u}_k) = W_H \cdot max(-\mathbf{u}_k,0)^2, \qquad if \ k \le N/2 $  \\  &  $c(\mathbf{u}_k) = W_L \cdot max(-\mathbf{u}_k,0)^2, \qquad else $\\\hline
		Defensive mode    & $c(\mathbf{u}_k) = W_H \cdot max(-\mathbf{u}_k,0)^2$\\\hline
		Cooperative mode   & $c(\mathbf{u}_k) = W_C \cdot (\mathbf{u}_k)^2$\\\hline
	\end{tabular}
\end{table}

\begin{figure*}[t]\centering
	\fontsize{8pt}{11pt}\selectfont
	\def\svgwidth{0.9\linewidth}
	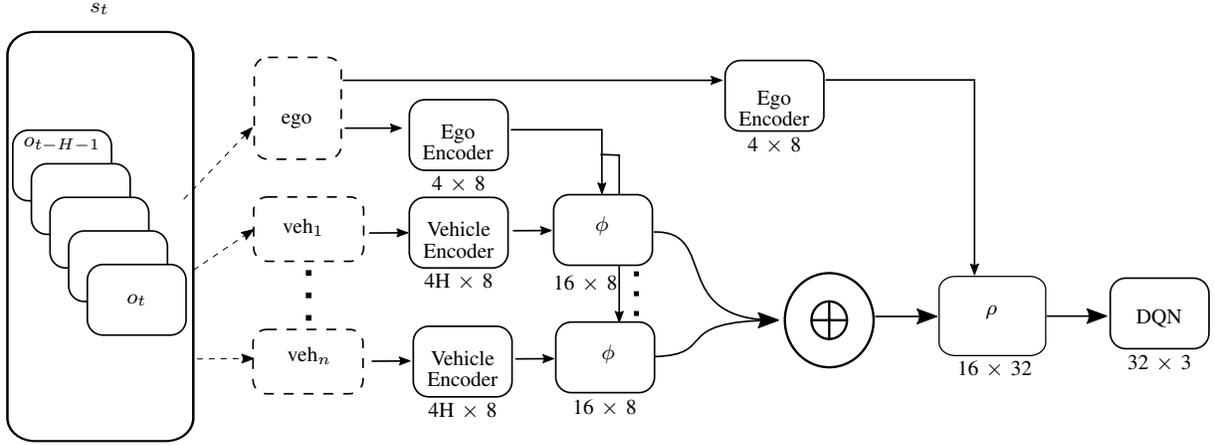
	\caption{Proposed Deep sets architecture including input and output dimensions of each learning block for extracting scalable features and learning action Q values for the RL agent.}\label{fig:architecture}
\end{figure*}

%%%%%%%%%%%%%%%%%%%%%%%%%%%%%%%%%%%%%%%%%%%%%%%%%%%%%%%%%%%%%%%%%%%%%%%%%%%%%%%%%%%%%%%%
\subsection{Learning Interactive Behavior with RL}
In this section we explain the proposed high level RL agent which selects suitable sub-maneuver from the MPC planner maneuver catalog according to the current observation and implicit predictions by the agent.

\subsubsection{Action Space}
Since the take-way maneuver is always selected if it is feasible, the action space of the RL agent consists of:
\begin{itemize}
\item Progressive give-way
\item Defensive give-way
\item Cooperative give-way
\end{itemize}
that identifies which of soft constraints related to the trajectory comfort cost defined in \ref{tab:give_ways_costs} should be applied for the safe give-way maneuver.

\subsubsection{Observation Space}
We assume that the distance and velocity of all surrounding vehicles are provided by a sensor fusion module and all relevant vehicles, i.e. those driving on the merging lane are extracted using a High Definition (HD)  map such as \cite{poggenhans_lanelet2} which provides infrastructure information about merging lanes and the conflict zones.
Using this information, we define the current observation as below:

\begin{equation}
o_t = \begin{tikzpicture}[baseline, decoration=brace]
\matrix (m) [matrix of math nodes,left delimiter=[,right delimiter={]^T}] {
	d_{\textrm{cs}} 		 & v_{\textrm{e}} &    d_{1} & ... & d_{n}  \\
	d_{\textrm{goal}} & a_{\textrm{e}} &    v_1 &  ... & v_n  \\
	%d_{\textrm{e},\textrm{goal}} & d_{\textrm{e},1} & ... & d_{\textrm{e},n} \\
};
\end{tikzpicture},
\label{eq:observation}
\end{equation}
where $d_{\textrm{cs}}$ is ego vehicle distance to the conflict zone, $d_{\textrm{goal}}$ its distance to the goal (after the merging area), $v_{\textrm{e}}$ its velocity and $a_{\textrm{e}}$ is its acceleration.
$d_i, v_i$ are distance and velocity of every vehicle relative to the conflict zone.
Finally, we provide a $k$-Markov approximation \cite{bouton2018k_markov} of the POMDP as input to the RL agent in order to enable $Q$-learning based on intention estimation of other vehicles from their observation history:
\begin{equation}
s_t = 
\begin{bmatrix}
o_t & o_{t-1} & ... & o_{t- (H-1)}
\end{bmatrix}
\end{equation}

\subsubsection{Reward Function}
For the reward function of the RL agent, we encourage fast and comfortable driving meaning that emergency full-stop maneuvers due to infeasible merging should be prohibited and also the policy should be less conservative and drive fast when the other vehicles are cooperative and probably reduce their velocity to create a merging gap for the ego vehicle.
Therefore, we propose this reward function:
\begin{equation}
\rm(o_t)=
\begin{cases}
$ 1$, & \text{if reach goal} \\
-\frac{\sum_{k=0}^{N_\text{p}} \text{max}(j_k - 5, 0)^2}{N_\text{p}} & \text{otherwise,} \\
\end{cases}
\end{equation}
where $j_k$ is the ego vehicle jerk at planning step $k$ and $N_\text{p}=6$ is the number of planning steps to execute one RL action.
By this rewarding scheme, the ego vehicle is motivated to reach the goal position as fast as possible (due to discount factor) and meanwhile it prevents to receive punishments caused by emergency jerk values.
In order to prevent emergency stopping maneuvers, the RL agent should implicitly predict the intention of other vehicles and only try to drive progressive when the other vehicles are cooperative or they are driving slowly behind the conflict zone.
%Selecting \textit{progressive give-way} maneuver makes a safe take-way maneuver feasible in the future (Figure \ref{fig:merging_mpc_options} red dashed line) and therefore the ego vehicle is able to enter into the conflict zone before other vehicles.
%In case the RL agent predicts non-cooperative behavior from other drivers and observers a risky situation for a future take-way maneuver, it selects \textit{defensive} or \textit{cooperative give-way} maneuver to slowly reduce the ego vehicle velocity and stop before the conflict zone. 

\subsubsection{DQN Architecture}
Based on Deep Q Networks approach \cite{dqn}, we design a neural network to learn future return of each RL action according to the input state.
One challenge here is that the number of surrounding vehicles considered in equation  \ref{eq:observation} which depends on the traffic density is unknown.
Therefore, we have to consider a big input size (16 vehicles in our experiments) to consider worst-case situation that may happen in dense traffic which can result in big DQN network and inefficient learning progress.
Utilizing \textit{Deep-Sets} mechanism \cite{zaheer2017deepsets, huegle2019dynamic}, we propose an efficient neural network architecture that can handle dynamic number of input elements and provide a fixed size feature vector for the main DQN network.
The overall structure of the proposed Deep-Sets and DQN networks are visible in Figure \ref{fig:architecture}.
In this graph, static information about the ego vehicle state and dynamic information about surrounding vehicles are compressed into smaller features using Encoder networks.
Then, each vehicle's feature vector is processed using $\phi$ network and the output of all processed features for all vehicles are combined using a permutation invariant operator (sum) in order to provide a fixed size feature vector for the DQN.
In the last layer behind the DQN, combined features about all vehicles is processed with the ego vehicle feature vector using $\rho$ network in order to provide all required information for the DQN agent.

%% file: figures/merging_mpc_options.pdf_tex
%% Accompanies image file 'merging_mpc_options.pdf' (pdf, eps, ps)
\begingroup%
  \makeatletter%
  \providecommand\color[2][]{%
    \errmessage{(Inkscape) Color is used for the text in Inkscape, but the package 'color.sty' is not loaded}%
    \renewcommand\color[2][]{}%
  }%
  \providecommand\transparent[1]{%
    \errmessage{(Inkscape) Transparency is used (non-zero) for the text in Inkscape, but the package 'transparent.sty' is not loaded}%
    \renewcommand\transparent[1]{}%
  }%
  \providecommand\rotatebox[2]{#2}%
  \newcommand*\fsize{\dimexpr\f@size pt\relax}%
  \newcommand*\lineheight[1]{\fontsize{\fsize}{#1\fsize}\selectfont}%
  \ifx\svgwidth\undefined%
    \setlength{\unitlength}{1306.15788978bp}%
    \ifx\svgscale\undefined%
      \relax%
    \else%
      \setlength{\unitlength}{\unitlength * \real{\svgscale}}%
    \fi%
  \else%
    \setlength{\unitlength}{\svgwidth}%
  \fi%
  \global\let\svgwidth\undefined%
  \global\let\svgscale\undefined%
  \makeatother%
  \begin{picture}(1,0.32266827)%
    \lineheight{1}%
    \setlength\tabcolsep{0pt}%
    \put(0.3854078,-0.53062919){\color[rgb]{0,0,0}\makebox(0,0)[lt]{\begin{minipage}{0.32483494\unitlength}\centering \end{minipage}}}%
    \put(0,0){\includegraphics[width=\unitlength,page=1]{merging_mpc_options.pdf}}%
    \put(0.72498688,0.31523676){\color[rgb]{0,0,0}\makebox(0,0)[lt]{\begin{minipage}{0.33631896\unitlength}\raggedright Other vehicle coopertative prediction\end{minipage}}}%
    \put(0.72517981,0.29265111){\color[rgb]{0,0,0}\makebox(0,0)[lt]{\begin{minipage}{0.3469828\unitlength}\raggedright Other vehicle worst-case prediction\end{minipage}}}%
    \put(0.72487663,0.26804241){\color[rgb]{0,0,0}\makebox(0,0)[lt]{\begin{minipage}{0.34780303\unitlength}\raggedright Ego vehicle defenstive give-way maneuver\end{minipage}}}%
    \put(0.72487663,0.24589457){\color[rgb]{0,0,0}\makebox(0,0)[lt]{\begin{minipage}{0.27774856\unitlength}\raggedright Ego vehicle progressive give-way maneuver\end{minipage}}}%
    \put(0,0){\includegraphics[width=\unitlength,page=2]{merging_mpc_options.pdf}}%
    \put(0.53039748,0.23435886){\color[rgb]{0,0,0}\makebox(0,0)[lt]{\begin{minipage}{0.13452759\unitlength}\centering \textit{d} / m\end{minipage}}}%
    \put(0.83064073,0.01404982){\color[rgb]{0,0,0}\makebox(0,0)[lt]{\begin{minipage}{0.13452759\unitlength}\centering \textit{t} / s\end{minipage}}}%
    \put(0.4764183,0.15687778){\color[rgb]{0,0,0}\makebox(0,0)[lt]{\begin{minipage}{0.13452759\unitlength}\centering conflict point\end{minipage}}}%
    \put(0,0){\includegraphics[width=\unitlength,page=3]{merging_mpc_options.pdf}}%
    \put(0.56518729,0.01944596){\color[rgb]{0,0,0}\makebox(0,0)[lt]{\begin{minipage}{0.13452759\unitlength}\centering $t_1$\end{minipage}}}%
    \put(0.53937225,0.01942373){\color[rgb]{0,0,0}\makebox(0,0)[lt]{\begin{minipage}{0.13452759\unitlength}\centering $t_0$\end{minipage}}}%
    \put(0,0){\includegraphics[width=\unitlength,page=4]{merging_mpc_options.pdf}}%
    \put(0.72498643,0.22504511){\color[rgb]{0,0,0}\makebox(0,0)[lt]{\begin{minipage}{0.3469828\unitlength}\raggedright Ego vehicle cooperative give-way maneuver\end{minipage}}}%
    \put(0,0){\includegraphics[width=\unitlength,page=5]{merging_mpc_options.pdf}}%
    \put(0.02655576,0.23918558){\color[rgb]{0,0,0}\makebox(0,0)[lt]{\begin{minipage}{0.13452759\unitlength}\centering \textit{d} / m\end{minipage}}}%
    \put(0.32679902,0.01887655){\color[rgb]{0,0,0}\makebox(0,0)[lt]{\begin{minipage}{0.13452759\unitlength}\centering \textit{t} / s\end{minipage}}}%
    \put(-0.02742341,0.16170451){\color[rgb]{0,0,0}\makebox(0,0)[lt]{\begin{minipage}{0.13452759\unitlength}\centering conflict point\end{minipage}}}%
    \put(0,0){\includegraphics[width=\unitlength,page=6]{merging_mpc_options.pdf}}%
    \put(0.06134557,0.02427269){\color[rgb]{0,0,0}\makebox(0,0)[lt]{\begin{minipage}{0.13452759\unitlength}\centering $t_1$\end{minipage}}}%
    \put(0.03553054,0.02425046){\color[rgb]{0,0,0}\makebox(0,0)[lt]{\begin{minipage}{0.13452759\unitlength}\centering $t_0$\end{minipage}}}%
    \put(0,0){\includegraphics[width=\unitlength,page=7]{merging_mpc_options.pdf}}%
    \put(0.21842209,0.29747784){\color[rgb]{0,0,0}\makebox(0,0)[lt]{\begin{minipage}{0.3469828\unitlength}\raggedright Other vehicle worst-case prediction\end{minipage}}}%
    \put(0,0){\includegraphics[width=\unitlength,page=8]{merging_mpc_options.pdf}}%
    \put(0.21814647,0.27925333){\color[rgb]{0,0,0}\makebox(0,0)[lt]{\begin{minipage}{0.3469828\unitlength}\raggedright Ego vehicle take-way maneuver\end{minipage}}}%
    \put(0,0){\includegraphics[width=\unitlength,page=9]{merging_mpc_options.pdf}}%
    \put(0.19547521,0.17119456){\color[rgb]{0,0,0}\makebox(0,0)[lt]{\begin{minipage}{0.08284931\unitlength}\centering $\Delta_t$\end{minipage}}}%
  \end{picture}%
\endgroup%

%% file: figures/dqn_arch.pdf_tex
%% Accompanies image file 'dqn_arch.pdf' (pdf, eps, ps)
\begingroup%
  \makeatletter%
  \providecommand\color[2][]{%
    \errmessage{(Inkscape) Color is used for the text in Inkscape, but the package 'color.sty' is not loaded}%
    \renewcommand\color[2][]{}%
  }%
  \providecommand\transparent[1]{%
    \errmessage{(Inkscape) Transparency is used (non-zero) for the text in Inkscape, but the package 'transparent.sty' is not loaded}%
    \renewcommand\transparent[1]{}%
  }%
  \providecommand\rotatebox[2]{#2}%
  \newcommand*\fsize{\dimexpr\f@size pt\relax}%
  \newcommand*\lineheight[1]{\fontsize{\fsize}{#1\fsize}\selectfont}%
  \ifx\svgwidth\undefined%
    \setlength{\unitlength}{1031.71690453bp}%
    \ifx\svgscale\undefined%
      \relax%
    \else%
      \setlength{\unitlength}{\unitlength * \real{\svgscale}}%
    \fi%
  \else%
    \setlength{\unitlength}{\svgwidth}%
  \fi%
  \global\let\svgwidth\undefined%
  \global\let\svgscale\undefined%
  \makeatother%
  \begin{picture}(1,0.36362156)%
    \lineheight{1}%
    \setlength\tabcolsep{0pt}%
    \put(0,0){\includegraphics[width=\unitlength,page=1]{dqn_arch.pdf}}%
    \put(0.76515987,0.11373253){\color[rgb]{0,0,0}\makebox(0,0)[lt]{\begin{minipage}{0.10696455\unitlength}\centering $\rho$\end{minipage}}}%
    \put(0,0){\includegraphics[width=\unitlength,page=2]{dqn_arch.pdf}}%
    \put(0.20551661,0.27252226){\color[rgb]{0,0,0}\makebox(0,0)[lt]{\begin{minipage}{0.07160749\unitlength}\centering ego\end{minipage}}}%
    \put(0.18952385,0.32687281){\color[rgb]{0,0,0}\makebox(0,0)[lt]{\begin{minipage}{0.07160752\unitlength}\centering \end{minipage}}}%
    \put(0.17643888,0.18615736){\color[rgb]{0,0,0}\makebox(0,0)[lt]{\begin{minipage}{0.14118641\unitlength}\centering $\text{veh}_1$\end{minipage}}}%
    \put(0.18136948,0.07927437){\color[rgb]{0,0,0}\makebox(0,0)[lt]{\begin{minipage}{0.14118641\unitlength}\centering $\text{veh}_n$\end{minipage}}}%
    \put(0,0){\includegraphics[width=\unitlength,page=3]{dqn_arch.pdf}}%
    \put(0.44672038,0.08086925){\color[rgb]{0,0,0}\makebox(0,0)[lt]{\begin{minipage}{0.10696455\unitlength}\centering $\phi$\end{minipage}}}%
    \put(0,0){\includegraphics[width=\unitlength,page=4]{dqn_arch.pdf}}%
    \put(0.03525874,0.36531721){\color[rgb]{0,0,0}\makebox(0,0)[lt]{\begin{minipage}{0.08482855\unitlength}\centering $s_t$\end{minipage}}}%
    \put(0,0){\includegraphics[width=\unitlength,page=5]{dqn_arch.pdf}}%
    \put(-0.00697279,0.25375684){\color[rgb]{0,0,0}\makebox(0,0)[lt]{\begin{minipage}{0.10696455\unitlength}\centering $o_{t-H-1}$\end{minipage}}}%
    \put(0,0){\includegraphics[width=\unitlength,page=6]{dqn_arch.pdf}}%
    \put(0.05498009,0.12315675){\color[rgb]{0,0,0}\makebox(0,0)[lt]{\begin{minipage}{0.10696455\unitlength}\centering $o_{t}$\end{minipage}}}%
    \put(0.33219387,0.13999436){\color[rgb]{0,0,0}\makebox(0,0)[lt]{\begin{minipage}{0.08765519\unitlength}\centering 4H $\times$ 8\end{minipage}}}%
    \put(0,0){\includegraphics[width=\unitlength,page=7]{dqn_arch.pdf}}%
    \put(0.33741333,0.18295231){\color[rgb]{0,0,0}\makebox(0,0)[lt]{\begin{minipage}{0.0770075\unitlength}\centering Vehicle\\  Encoder\end{minipage}}}%
    \put(0.33551392,0.03283908){\color[rgb]{0,0,0}\makebox(0,0)[lt]{\begin{minipage}{0.08765519\unitlength}\centering 4H $\times$ 8\end{minipage}}}%
    \put(0,0){\includegraphics[width=\unitlength,page=8]{dqn_arch.pdf}}%
    \put(0.34073338,0.07579703){\color[rgb]{0,0,0}\makebox(0,0)[lt]{\begin{minipage}{0.0770075\unitlength}\centering Vehicle \\ Encoder\end{minipage}}}%
    \put(0.33222423,0.22120793){\color[rgb]{0,0,0}\makebox(0,0)[lt]{\begin{minipage}{0.08765519\unitlength}\centering 4 $\times$ 8\end{minipage}}}%
    \put(0,0){\includegraphics[width=\unitlength,page=9]{dqn_arch.pdf}}%
    \put(0.33709347,0.26428514){\color[rgb]{0,0,0}\makebox(0,0)[lt]{\begin{minipage}{0.07664034\unitlength}\centering Ego \\ Encoder\end{minipage}}}%
    \put(0.43736528,0.13854014){\color[rgb]{0,0,0}\makebox(0,0)[lt]{\begin{minipage}{0.08765519\unitlength}\centering 16 $\times$ 8\end{minipage}}}%
    \put(0.45333459,0.03555121){\color[rgb]{0,0,0}\makebox(0,0)[lt]{\begin{minipage}{0.08765519\unitlength}\centering 16 $\times$ 8\end{minipage}}}%
    \put(0.77745998,0.06723737){\color[rgb]{0,0,0}\makebox(0,0)[lt]{\begin{minipage}{0.08765519\unitlength}\centering 16 $\times$   32\end{minipage}}}%
    \put(0,0){\includegraphics[width=\unitlength,page=10]{dqn_arch.pdf}}%
    \put(0.59458134,0.25383404){\color[rgb]{0,0,0}\makebox(0,0)[lt]{\begin{minipage}{0.08765519\unitlength}\centering 4 $\times$ 8\end{minipage}}}%
    \put(0,0){\includegraphics[width=\unitlength,page=11]{dqn_arch.pdf}}%
    \put(0.59945058,0.29194569){\color[rgb]{0,0,0}\makebox(0,0)[lt]{\begin{minipage}{0.0770075\unitlength}\centering Ego Encoder\end{minipage}}}%
    \put(0.91454306,0.07285974){\color[rgb]{0,0,0}\makebox(0,0)[lt]{\begin{minipage}{0.08765519\unitlength}\centering 32 $\times$ 3\end{minipage}}}%
    \put(0,0){\includegraphics[width=\unitlength,page=12]{dqn_arch.pdf}}%
    \put(0.91941235,0.11097139){\color[rgb]{0,0,0}\makebox(0,0)[lt]{\begin{minipage}{0.0770075\unitlength}\centering DQN\end{minipage}}}%
    \put(0,0){\includegraphics[width=\unitlength,page=13]{dqn_arch.pdf}}%
    \put(0.44146861,0.18780839){\color[rgb]{0,0,0}\makebox(0,0)[lt]{\begin{minipage}{0.10696455\unitlength}\centering $\phi$\end{minipage}}}%
    \put(0,0){\includegraphics[width=\unitlength,page=14]{dqn_arch.pdf}}%
  \end{picture}%
\endgroup%

%% file: content/04_results_and_evaluation.tex
\section{Results and Evaluation}\label{sec:evaluations}
\subsection{Simulation Environment}
For training and evaluating the proposed RL agent, we generate random scenarios including random number of vehicles based on vehicle insertion probability ($p_\text{new}$) which tunes the traffic density of simulations and random desired velocities for each vehicle.
All of the vehicles use Intelligent Driver Model \cite{treiber2000IDM} with maximum acceleration  $2\mpsq$, minimum deceleration $-10\mpsq$ and comfortable deceleration $1.6\mpsq$ to follow their desired velocity and prevent collisions with front vehicles with safety distance $2.0 \text{m}$ and time headway $2.0 \text{s}$. 
The IDM, however, does not react to the ego vehicle and therefore a vehicle can have collision with the ego vehicle.
The desired velocity for each vehicle is selected randomly by normal distribution with a randomly selected $\mu$ from \{5, 10, 15$\mps$\}  and $\sigma$=2$\mps$.
In order to simulate cooperative drivers, vehicles are with probability $p_{\text{coop}}$=\{0.0, 0.2, ..., 1.\} cooperative and yield to the ego vehicle by setting their desired velocity to zero when the ego vehicle is close to the conflict zone.

For training the DQN agent, we used double DQN \cite{double_dqn} implementation with target network update frequency of 200, learning rate=9e-7 with Adam optimizer \cite{kingma2014adam}, $\gamma$=0.99 and applied $\epsilon$-greedy policy exploration with $\epsilon_\text{init}$=0.3 and $\epsilon_\text{final}$=0.2. 
RL actions were applied every 600 ms and low level planner every 100 ms.

\begin{figure*}
	\centering
	\begin{minipage}{.45\textwidth}
\centering
\includegraphics[width=0.8\linewidth]
{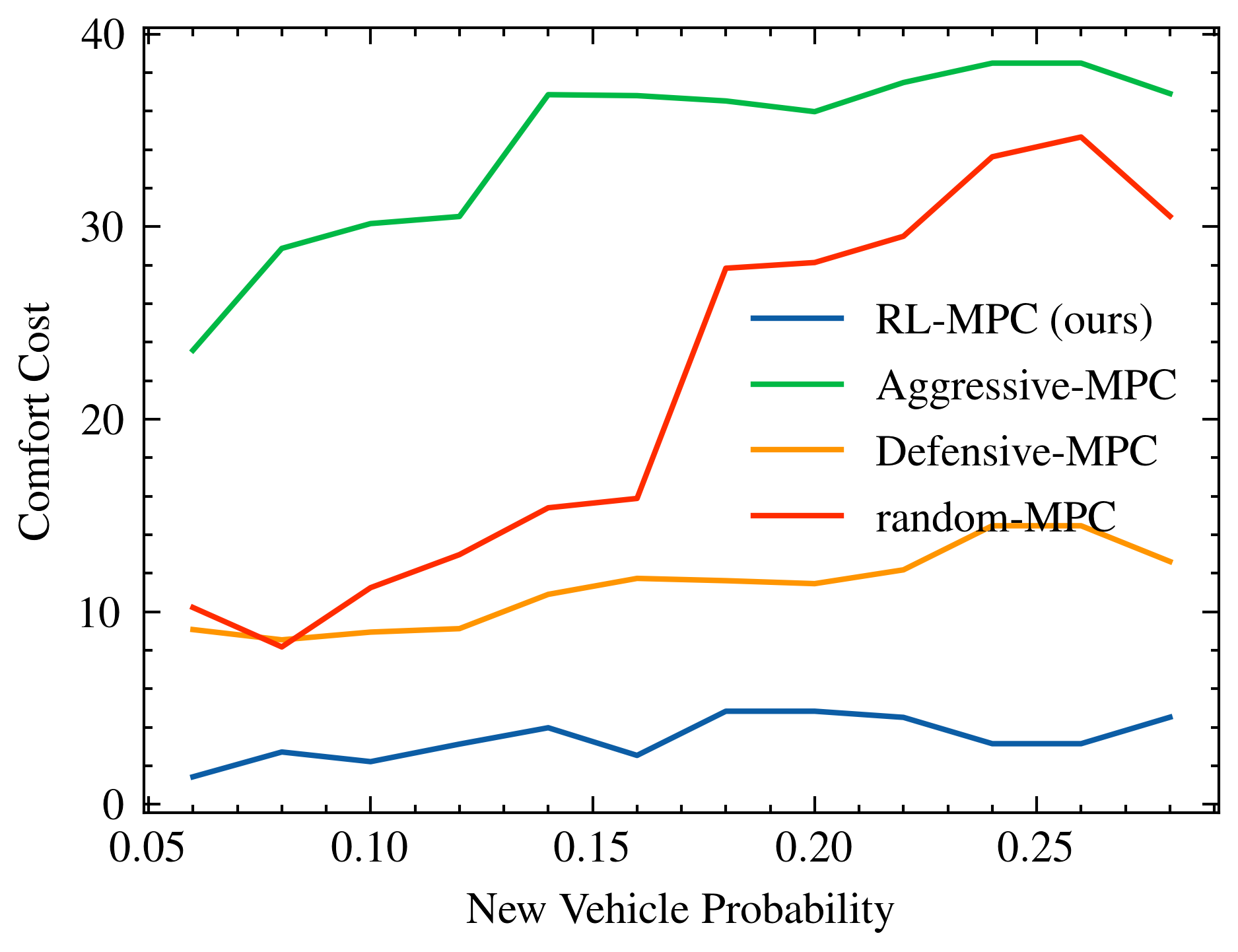}
\caption{Average comfort cost for each policy with different traffic densities.}\label{fig:emg_insert}

	\end{minipage}%
\hspace{.05\textwidth}	
	\begin{minipage}{.45\textwidth}
\centering
\includegraphics[width=0.8\linewidth]
{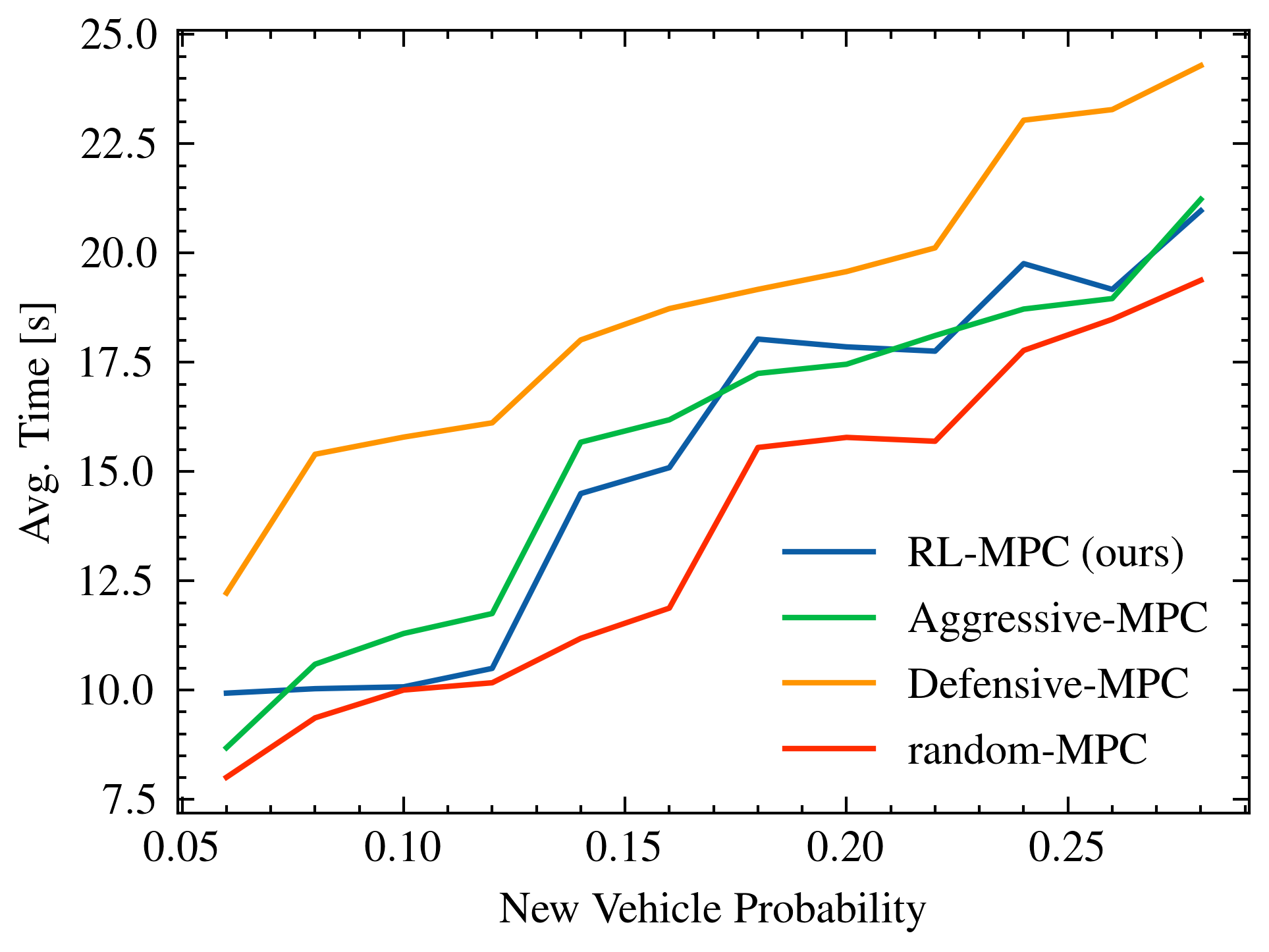}
\caption{Average crossing time for each policy with different traffic densities.}\label{fig:time_insert}

	\end{minipage}
\end{figure*}

\subsection{Effect of the Observation History Length}
In order to study the impact of observation history on the efficiency of the learned policy, we trained different DQN agents with multiple observation history lengths ($H$) and compared their efficiency in Table \ref{table:score_history}.
We found out that by reducing the history length, the learned policy becomes slower in average due to limited information about vehicles state in the past which prevents the agent to efficiently identify cooperative drivers and therefore it behaves more conservatively.
With higher history lengths like 30 and 36, the agents scarifies comfort to drive faster.
We used history of $H$=24 ($\times$100 ms) in our experiments which shows the best trade-off in Table \ref{table:score_history}.

\begin{table}[t]
	\centering
	\caption{Comparing efficiency of different history lengths. }\label{tab:score_history}
	\begin{tabular}{l|l|l|l|l|l|l}
		History ($\times$100 ms) & 1		& 6     & 12   		& 24    		& 30    	& 36    \\ \Xhline{3\arrayrulewidth}
		Avg. Time (s)           & 15.9		& 15.9 & 15.3 	& 14.1 			& 13.5 	& \textbf{12.9} \\ \hline
		Avg. Discomfort         & 7.45  	& 1.63  & 1.08  	& \textbf{1.08}	& 2.16  	& 2.19  \\ \hline
		Total Cost $C^\pi$                   & 1885 	& 412.6 & 253 	& \textbf{211} & 395 	& 367
	\end{tabular}\label{table:score_history}
\end{table}

\subsection{Comparison with Baseline Agents}
We designed different types of MPC agents as baseline policies and compare their efficiency with the proposed RL base agent.
\textit{Progressive-MPC} and \textit{Defensive-MPC} agents consider merely progressive and defensive give-way maneuvers respectively.
\textit{Random-MPC} selects one of the give-way sub-maneuvers randomly.
We also design a \textit{Neutral-MPC} which uses a normal comfort cost function that uniformly punishes all jerk control commands to drive as comfortable as possible.
It should be noted that all of the designed policies including the proposed RL-MPC agent are safe due to the worst-case safety verification constraints applied in optimization. 

We compare efficiency of baseline policies with 50 benchmark episodes.
We run these episodes in several iterations with different traffic densities in the simulation configuration. 
For each policy, we measured average crossing time and also average comfort cost as:
\begin{equation}\label{eq:je}
J_\text{emg}^{\pi} = \frac{\sum_{k=0}^{N_\text{e}} \text{max}(j_k - 5, 0)^2}{N_\text{e}},
\end{equation}
where $j_k$ is the ego vehicle jerk applied at planning time step $k$ based on sub-maneuvers selected by policy $\pi$ during driving and $N_\text{e}$ is the total number of planning steps.
We assume jerk values below 5 $\mpsc$ as comfortable jerk commands and filter them out in this cost function.
The squared values of $j_k$ helps to give more cost to policies with a few big emergency jerks comparing to policies with numerous small emergency jerks.

\begin{figure}
	\centering
	\includegraphics[width=0.8\linewidth]
	{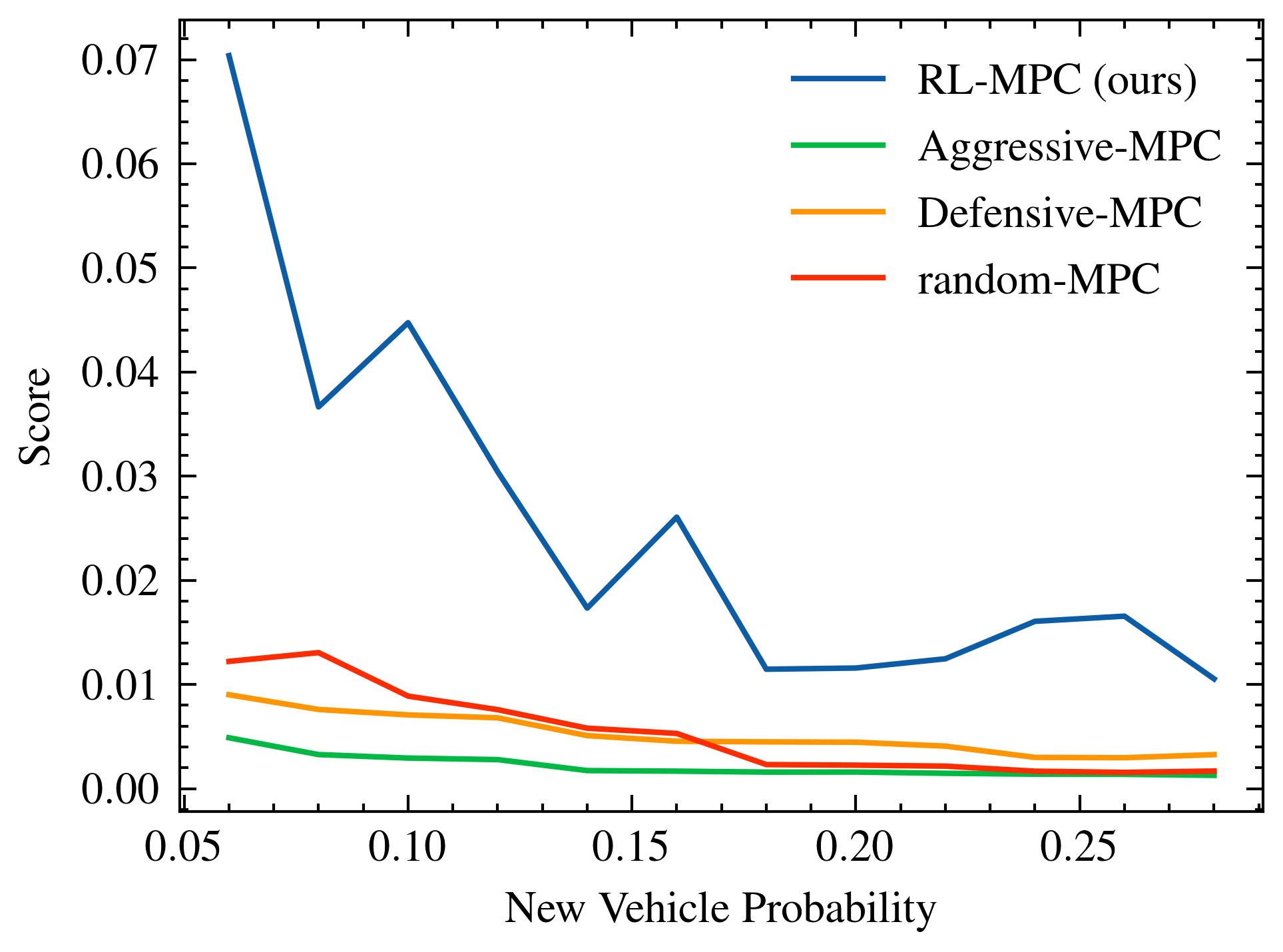}
	\caption{Average return of each policy with different traffic densities.}
	\label{fig:score_insert}
	
\end{figure}

Figures  \ref{fig:emg_insert} and \ref{fig:time_insert} depict average comfort cost and average time for different policies and different traffic densities.
As it is visible, all policies become slower for higher traffic densities (Figure \ref{fig:time_insert}), and the random-MPC is the fastest one.
The defensive-MPC is the slowest policy since it conservatively selects defensive give-way maneuver in all scenarios.
According to Figure \ref{fig:emg_insert}, the RL-MPC has the lowest amount of emergency jerks resulting in more comfortable maneuvers.
Moreover, RL-MPC has highest average return in figure  \ref{fig:score_insert}.
The main improvement of RL-MPC as a result of its learning ability for better scene understanding is visible in figures  \ref{fig:emg_insert} and \ref{fig:score_insert} as the only policy which improves both speed and comfort of maneuvers at the same time and receives the highest reward.

% \usepackage{multirow}
\begin{table*}[ht]
	\centering
	\caption{Comparing all policies with different cooperation and speed levels for other drivers in simulation.}\label{table:all_configurations}
	\begin{tabular}{l|l|l|lllll}
		\Xhline{3\arrayrulewidth}
		\multicolumn{2}{l|}{Env. Config.}                   & \multirow{2}{*}{Metric} & \multirow{2}{*}{Random-MPC} & \multirow{2}{*}{Progressive-MPC} & \multirow{2}{*}{Neutral-MPC} & \multirow{2}{*}{Defensive-MPC} & \multirow{2}{*}{RL-MPC (ours)} \\ \cline{1-2}
		Avg. Vel.                & $p_\text{coop}$               &                         &                             &                                 &                              &                                &                                \\ \Xhline{3\arrayrulewidth}
		\multirow{6}{*}{8 $\mps$}  
		& \multirow{3}{*}{0.1} & Avg. Time               & 11.2                        & \textbf{11.0}                   & 13.8                         & 14.7                           & 13.3                           \\ %\cline{3-8} 
		&                      & Comfort Cost            & 13.5                        & 27.9                            & 1.9                          & 8.8                            & \textbf{0.9}                            \\ %\cline{3-8} 
		&                      & Total Cost $C^\pi$      & 1693                       & 3375                             & {418}                       & 1901                            & \textbf{159}                            \\ \cline{2-8} 
		%%%%%%%%%%%%%%%%%%%%%%%%%%%%%%%%%%%%%%%%%%%%%%%%%%%%%%%%%%%%%%%%%%%%%%%%%%%%%%%%%%%%%%%%%%%%%%%%%%%%%%%%%%%%%%%%%%%%%%%%%%%%%%%%%%%%%%%%%%%%%%%%%%%%%%%%%%%%%%%%%%%%%%%%%%%%%%%%%%%%%%%%%%%%%%%%%%%%%%%
		& \multirow{3}{*}{0.7} & Avg. Time               & \textbf{9.1}                & 9.2                            & 12.8                         & 12.2                           & 9.4                           \\ %\cline{3-8} 
		&                      & Comfort Cost            & 13.8                        & 26.9                            & \textbf{1.4}                 & 7.6                           & 1.7                            \\ %\cline{3-8} 
		&                      & Total Cost $C^\pi$      & 1142                         & 2276                             & 229               & 1131                            & \textbf{150}                           \\ \Xhline{3\arrayrulewidth}
		\multirow{6}{*}{15 $\mps$}
		%%%%%%%%%%%%%%%%%%%%%%%%%%%%%%%%%%%%%%%%%%%%%%%%%%%%%%%%%%%%%%%%%%%%%%%%%%%%%%%%%%%%%%%%%%%%%%%%%%%%%%%%%%%%%%%%%%%%%%%%%%%%%%%%%%%%%%%%%%%%%%%%%%%%%%%%%%%%%%%%%%%%%%%%%%%%%%%%%%%%%%%%%%%%%%%%%%%%%%%
		& \multirow{3}{*}{0.3} & Avg. Time               & \textbf{11.2}              & 11.4                            & 14.1                         & 14.4                           & 13.2                           \\ %\cline{3-8} 
		&                      & Comfort Cost             & 21.2                        & 31.7                            & 1.4                          & 7.4                            & \textbf{0.9}                            \\ %\cline{3-8} 
		&                      & Total Cost $C^\pi$       & 2659                         & 4119                           & 278                          & 1534                            & \textbf{156}                            \\ \cline{2-8} 
		%%%%%%%%%%%%%%%%%%%%%%%%%%%%%%%%%%%%%%%%%%%%%%%%%%%%%%%%%%%%%%%%%%%%%%%%%%%%%%%%%%%%%%%%%%%%%%%%%%%%%%%%%%%%%%%%%%%%%%%%%%%%%%%%%%%%%%%%%%%%%%%%%%%%%%%%%%%%%%%%%%%%%%%%%%%%%%%%%%%%%%%%%%%%%%%%%%%%%%%
		& \multirow{3}{*}{0.7} & Avg. Time               & \textbf{8.2}                & 9.5                             & 11.7                         & 11.7                           & 9.7                            \\ %\cline{3-8} 
		&                      & Comfort Cost            & 19.9                        & 28.9                            & 2.2                          & 3.7                            & \textbf{1.1}                            \\ %\cline{3-8} 
		&                      & Total Cost $C^\pi$      & 1338                         & 2608                             & 301                          & 506                            & \textbf{103}                            \\ \Xhline{3\arrayrulewidth}
	\end{tabular}
\end{table*}

\subsection{Evaluation with Different Cooperation Levels}
In order to evaluate efficiency of the learned policy more specifically, we compared them in different experiments with different cooperation levels and average velocities for other vehicles.
We also created a new metric for each policy called \textit{total cost} which helps to compare policies in both terms of comfort and speed together:
\begin{equation}
C^\pi = J_\text{emg}^\pi \times (T^\pi)^2,
\end{equation}
where $J_\text{emg}^\pi$ is computed based on equation \ref{eq:je} and $T^\pi$ is the average time needed to drive in one episode using policy $\pi$ in evaluation episodes.
The average time, comfort cost and total cost for each policy are provided in Table \ref{table:all_configurations}.
We divided the experiments into two major categories: low speed tests with average velocity of 8$\mps$ and high speed tests with 15$\mps$ average velocity for other vehicles.
For each category we evaluated two different cooperation levels for other drivers to simulate traffic situations with low or high amount of cooperativeness.
According to Table \ref{table:all_configurations}, proposed RL-MPC agent is the only policy which improves both comfort and speed of drivings in different configurations at the same time.
The Neutral-MPC agent has also low comfort cost but it is slower than RL-MPC since it has no reaction to cooperative drivers.
On the other hand, Random-MPC and Progressive-MPC are faster than others but they both have high comfort cost.
Therefore, the RL-MPC agent has the lowest total cost in all configurations.

%% file: content/05_conclusions.tex
\section{Conclusions and Future Work}\label{sec:conclusions}
In this paper we presented an RL trajectory planing pipeline for safe and comfortable automated driving in merging scenarios.
Instead of low level control commands, the proposed RL algorithm decides about safe sub-maneuvers which are optimized using a low level trajectory planner.
Therefore, the RL agent is not responsible for safety, but it tries to maximize comfort and speed of maneuvers.
In order to provide a scalable policy, we use Deep-Sets scheme which helps to process dynamic number of input elements (surrounding vehicles) efficiently and provide fixed size features for the DQN neural network.
Using the history of vehicles states, the proposed RL agent can implicitly identify cooperative drivers and generate suitable actions to comfortably merge into the other lanes in our simulation benchmarks. 
On the other hand, thanks to the safety constraints utilized inside the low level planner, the policy has no collision in all evaluation experiments.
For future works, we would like to evaluate efficiency of the proposed approach with real data and implement it on our experimental vehicle.

%% file: 06_acknowledgements.tex
\section*{acknowledgement}
The authors would like to thank Maximilian Naumann for the fruitful discussions on the way to this work. 
This research is accomplished within the project ``UNICARagil'' (FKZ 6EMO0287).
We acknowledge the financial support for the project by the Federal Ministry of Education and Research of Germany (BMBF).

%% file: references.bib
@article{safe_multi_agent,
	title={Safe, multi-agent, reinforcement learning for autonomous driving},
	author={Shalev-Shwartz, Shai and Shammah, Shaked and Shashua, Amnon},
	journal={arXiv preprint arXiv:1610.03295},
	year={2016}
}

@INPROCEEDINGS{isele_navigating,
	author={D. {Isele} and R. {Rahimi} and A. {Cosgun} and K. {Subramanian} and K. {Fujimura}},
	booktitle={2018 IEEE International Conference on Robotics and Automation (ICRA)},
	title={Navigating Occluded Intersections with Autonomous Vehicles Using Deep Reinforcement Learning},
	year={2018},
	volume={},
	number={},
	pages={2034-2039},
	doi={10.1109/ICRA.2018.8461233},
	ISSN={},
	month={May},}

@inproceedings{tram2019rl_mpc,
	title={Learning when to drive in intersections by combining reinforcement learning and model predictive control},
	author={Tram, Tommy and Batkovic, Ivo and Ali, Mohammad and Sj{\"o}berg, Jonas},
	booktitle={2019 IEEE Intelligent Transportation Systems Conference (ITSC)},
	pages={3263--3268},
	year={2019},
	organization={IEEE}
}

@article{kamran2020risk,
	title={Risk-Aware High-level Decisions for Automated Driving at Occluded Intersections with Reinforcement Learning},
	author={Kamran, Danial and Lopez, Carlos Fernandez and Lauer, Martin and Stiller, Christoph},
	journal={arXiv preprint arXiv:2004.04450},
	year={2020}
}

@inproceedings{bouton2019cooperation,
	title={Cooperation-aware reinforcement learning for merging in dense traffic},
	author={Bouton, Maxime and Nakhaei, Alireza and Fujimura, Kikuo and Kochenderfer, Mykel J},
	booktitle={2019 IEEE Intelligent Transportation Systems Conference (ITSC)},
	pages={3441--3447},
	year={2019},
	organization={IEEE}
}

@article{althoff2014online,
	title={Online verification of automated road vehicles using reachability analysis},
	author={Althoff, Matthias and Dolan, John M},
	journal={IEEE Transactions on Robotics},
	volume={30},
	number={4},
	pages={903--918},
	year={2014},
	publisher={IEEE}
}

@inproceedings{zaheer2017deepsets,
	title={Deep sets},
	author={Zaheer, Manzil and Kottur, Satwik and Ravanbakhsh, Siamak and Poczos, Barnabas and Salakhutdinov, Russ R and Smola, Alexander J},
	booktitle={Advances in neural information processing systems},
	pages={3391--3401},
	year={2017}
}

@inproceedings{huegle2019dynamic,
	title={Dynamic Input for Deep Reinforcement Learning in Autonomous Driving},
	author={Huegle, Maria and Kalweit, Gabriel and Mirchevska, Branka and Werling, Moritz and Boedecker, Joschka},
	booktitle={2019 IEEE/RSJ International Conference on Intelligent Robots and Systems (IROS)},
	pages={7566--7573},
	year={2019},
	organization={IEEE}
}

@inproceedings{muller2019risk,
	title={A risk and comfort optimizing motion planning scheme for merging scenarios},
	author={M{\"u}ller, Johannes and Buchholz, Michael},
	booktitle={2019 IEEE Intelligent Transportation Systems Conference (ITSC)},
	pages={3155--3161},
	year={2019},
	organization={IEEE}
}

@inproceedings{batkovic2019mpc_prediction,
	title={Real-time constrained trajectory planning and vehicle control for proactive autonomous driving with road users},
	author={Batkovic, Ivo and Zanon, Mario and Ali, Mohammad and Falcone, Paolo},
	booktitle={2019 18th European Control Conference (ECC)},
	pages={256--262},
	year={2019},
	organization={IEEE}
}

@inproceedings{hubmann2018merging,
	title={A belief state planner for interactive merge maneuvers in congested traffic},
	author={Hubmann, Constantin and Schulz, Jens and Xu, Gavin and Althoff, Daniel and Stiller, Christoph},
	booktitle={2018 21st International Conference on Intelligent Transportation Systems (ITSC)},
	pages={1617--1624},
	year={2018},
	organization={IEEE}
}

@inproceedings{bouton2018k_markov,
	title={Utility decomposition with deep corrections for scalable planning under uncertainty},
	author={Bouton, Maxime and Julian, Kyle and Nakhaei, Alireza and Fujimura, Kikuo and Kochenderfer, Mykel J},
	booktitle={Proceedings of the 17th International Conference on Autonomous Agents and MultiAgent Systems},
	pages={462--469},
	year={2018}
}

@article{bellman_dynamic,
	title={Dynamic programming},
	author={Bellman, Richard},
	journal={Science},
	volume={153},
	number={3731},
	pages={34--37},
	year={1966},
	publisher={American Association for the Advancement of Science}
}

@article{qlearning,
	author = {Watkins, Christopher J C H and Dayan, Peter},
	doi = {10.1007/BF00992698},
	journal = {Machine Learning},
	number = {3},
	pages = {279--292},
	title = {{Q-learning}},
	url = {https://doi.org/10.1007/BF00992698},
	volume = {8},
	year = {1992}
}

@book{sutton_barto_rl,
	title = {Reinforcement Learning: An Introduction},
	shorttitle = {Reinforcement Learning},
	author = {Sutton, Richard S. and Barto, Andrew},
	year = {2018},
	publisher = {{The MIT Press}}
}

@article{dqn,
	
	archivePrefix = {arXiv},
	arxivId = {1312.5602},
	author = {Mnih, Volodymyr and Silver, David},
	eprint = {1312.5602},
	pmid = {24966830},
	title = {{Playing Atari with Deep Reinforcement Learning}},
	url = {http://arxiv.org/abs/1509.02971},
	year = {2013}
	
	
}

@inproceedings{double_dqn,
	title={Deep reinforcement learning with double q-learning},
	author={Van Hasselt, Hado and Guez, Arthur and Silver, David},
	booktitle={Thirtieth AAAI conference on artificial intelligence},
	year={2016}
}

@article{kingma2014adam,
	title={Adam: A method for stochastic optimization},
	author={Kingma, Diederik P and Ba, Jimmy},
	journal={arXiv preprint arXiv:1412.6980},
	year={2014}
}

@inproceedings{poggenhans_lanelet2,
	title={Lanelet2: A high-definition map framework for the future of automated driving},
	author={Poggenhans, Fabian and Pauls, Jan-Hendrik and Janosovits, Johannes and Orf, Stefan and Naumann, Maximilian and Kuhnt, Florian and Mayr, Matthias},
	booktitle={2018 21st International Conference on Intelligent Transportation Systems (ITSC)},
	pages={1672--1679},
	year={2018},
	organization={IEEE}
}

@article{treiber2000IDM,
	title={Congested traffic states in empirical observations and microscopic simulations},
	author={Treiber, Martin and Hennecke, Ansgar and Helbing, Dirk},
	journal={Physical review E},
	volume={62},
	number={2},
	pages={1805},
	year={2000},
	publisher={APS}
}
